\documentclass[english]{article}
\usepackage{graphicx} 
\usepackage[most]{tcolorbox}
\usepackage{graphicx} 
\usepackage{subcaption}
\usepackage{wrapfig}
\usepackage{xcolor}
\usepackage{xspace}
\usepackage{booktabs}
\usepackage{url}
\usepackage{multirow}
\usepackage[linesnumbered, boxed]{algorithm2e}
\usepackage{amsmath, amsthm}
\usepackage{listings}

\usepackage[utf8]{inputenc} 
\usepackage[T1]{fontenc}    
\usepackage{hyperref}       
\usepackage{url}            
\usepackage{amsfonts}       
\usepackage{nicefrac}       
\usepackage{microtype}      

\usepackage{amssymb}
\usepackage{mylstlinebgrd}

\usepackage{geometry}
\usepackage[numbers, sort]{natbib} 
\usepackage{babel}
\usepackage{siunitx}
\usepackage{bm}
\usepackage{bbm}
\usepackage{caption}
\usepackage{appendix}

\definecolor{codegreen}{rgb}{0,0.6,0}
\definecolor{codegray}{rgb}{0.5,0.5,0.5}
\definecolor{codepurple}{rgb}{0.58,0,0.82}
\definecolor{backcolour}{rgb}{0.95,0.95,0.92}

\definecolor{eclipseBlue}{RGB}{42,0.0,255}
\definecolor{eclipseGreen}{RGB}{63,180,95}
\definecolor{eclipsePurple}{RGB}{175,0,25}
\definecolor{codewhite}{rgb}{0.70,0.70,0.70}

\definecolor{moderator}{HTML}{e3716e}
\definecolor{player1}{HTML}{bdb5e1}
\definecolor{player2}{HTML}{7ac7e2}
\definecolor{player3}{HTML}{f7df87}
\definecolor{player4}{HTML}{54beaa}
\definecolor{player5}{HTML}{2983b1}
\definecolor{player6}{HTML}{eca680}

\lstdefinestyle{mystyle}{
    backgroundcolor=\color{backcolour},   
    commentstyle=\color{codegreen},
    keywordstyle=\color{magenta},
    numberstyle=\tiny\color{codegray},
    stringstyle=\color{codepurple},
    basicstyle=\ttfamily\footnotesize,
    breakatwhitespace=false,         
    breaklines=true,
    breakindent=-3.5pt,
    captionpos=b,                    
    keepspaces=true,                 
    numbers=left,                    
    numbersep=5pt,                  
    showspaces=false,                
    showstringspaces=false,
    showtabs=false,                  
    tabsize=4,
    postbreak={
    	\mbox{
    		\lst@linebreakbgrd
    		\rotatebox[y=0.7ex]{180}{\color{black}$\Lsh\,$}
    	}
    },
}

\lstdefinelanguage{Dialog}{
	backgroundcolor=\color{backcolour},   
	keywordstyle=\color{magenta},
	numberstyle=\tiny\color{codegray},
	basicstyle=\ttfamily\footnotesize,
	breakatwhitespace=false,         
	breaklines=true,   
    breakindent=-3.5pt,
	captionpos=b,                    
	keepspaces=true,                 
	numbers=left,                    
	numbersep=5pt,                  
	showspaces=false,                
	showstringspaces=false,
	showtabs=false,                  
	tabsize=4,
	morecomment = [l][\color{eclipseGreen}\bfseries]{Assistant:},
    morecomment = [s][\color{eclipseGreen}\bfseries]{1.}{more.},
    morecomment = [l][\color{eclipseBlue}\bfseries]{User:},
}

\lstdefinelanguage{Werewolf}{
	backgroundcolor=\color{backcolour},   
	keywordstyle=\color{magenta},
	numberstyle=\tiny\color{codegray},
	basicstyle=\ttfamily\footnotesize,
	breakatwhitespace=false,         
	breaklines=true,   
    breakindent=-3.5pt,
	captionpos=b,                    
	keepspaces=true,                 
	numbers=left,                    
	numbersep=5pt,                  
	showspaces=false,                
	showstringspaces=false,
	showtabs=false,                  
	tabsize=4,
    morecomment = [l][\color{player1}\bfseries]{Player1:},
     morecomment = [l][\color{player2}\bfseries]{Player2 :},
     morecomment = [l][\color{player3}\bfseries]{Player3 :},
     morecomment = [l][\color{player4}\bfseries]{Player4 :},
     morecomment = [l][\color{player5}\bfseries]{Player5 :},
     morecomment = [l][\color{player6}\bfseries]{Player6 :},
     morecomment = [s][\color{moderator}\bfseries]{Moderator :}{\}},
}

\newcommand{\ours}{\texttt{KIMAs}\xspace}

\newcommand{\mypar}[1]{\noindent{\textbf{#1}}}
\newcommand{\mysubpar}[1]{\noindent{\textit{#1}}}

\title{\ours: A Configurable Knowledge Integrated  Multi-Agent  System}
\date{}

\author{
Zitao Li\thanks{Co-first authors.}, Fei Wei\footnotemark[1], Yuexiang Xie\footnotemark[1], Dawei Gao, Weirui Kuang, Zhijian Ma, Bingchen Qian, 
\\Yaliang Li, Bolin Ding
\\
\\
\small{Alibaba Group}
}

\begin{document}

\maketitle
\begin{abstract}
Knowledge-intensive conversations supported by large language models (LLMs) have become one of the most popular and helpful applications that can assist people in different aspects.
Many current knowledge-intensive applications are centered on retrieval-augmented generation (RAG) techniques.
While many open-source RAG frameworks facilitate the development of RAG-based applications, they often fall short in handling practical scenarios complicated by heterogeneous data in topics and formats,  conversational context management, and the requirement of low-latency response times. 
This technical report presents a configurable \underline{k}nowledge \underline{i}ntegrated \underline{m}ulti-\underline{a}gent \underline{s}ystem, \ours, to address these challenges.   
\ours features a flexible and configurable system for integrating diverse knowledge sources with 1) context management and query rewrite mechanisms to improve retrieval accuracy and multi-turn conversational coherency, 2) efficient knowledge routing and retrieval, 3) simple but effective filter and reference generation mechanisms, and 4) optimized parallelizable multi-agent pipeline execution.
Our work provides a scalable framework for advancing the deployment of LLMs in real-world settings.
To show how \ours can help developers build knowledge-intensive applications with different scales and emphases, we demonstrate how we configure the system to three applications already running in practice with reliable performance.\footnote{The source code is under review and will be available soon.}
\end{abstract}

\section{Introduction}
Large language models (LLMs) have had a profound impact on various aspects of people's lives, particularly as the foundational technology behind conversational applications such as chatbots. 
These models have become indispensable as virtual assistants, offering powerful capabilities for various tasks, including addressing common-sense queries, generating summaries for academic papers~\cite{lin-etal-2024-arxiv}, and solving programming challenges and tasks~\cite{jimenezswe}.  
Despite their impressive functionality, LLMs are of some limitations. 
Issues such as hallucinations and the inability to provide the most up-to-date information or private knowledge hinder their reliability in directly serving for knowledge-intensive applications. 
These shortcomings can be mitigated by integrating LLMs with external information in the input context~\cite{min2022rethinking, wei2022emergent}.
One notable approach is retrieval-augmented generation (RAG) techniques~\cite{izacard2020leveraging, borgeaud2022improving}, which enhances LLMs by equipping them with retrieval capabilities, allows LLMs to address questions that exceed the scope of their pre-trained internal knowledge. 
RAG has proven highly effective in improving performance on question-answering (QA) tasks emphasizing faithfulness to truths, showcasing its potential to bridge the gap between static pre-trained knowledge and dynamic, context-specific information.

While many real-world applications have adopted RAG techniques~\cite{perplexity, kimi}, open-source frameworks have also emerged to facilitate the adaptation of RAG to a wide range of tasks~\cite{llamaindex, langchain} for the public to hold RAG application services themselves with local data. 
While these open-source RAG frameworks provide convenient starting points for building RAG-based applications, there remain significant opportunities for improvement, especially in more practical and complicated scenarios, e.g., efficient multi-source knowledge retrieval, which provides primary motivations for this paper.

\mypar{Challenge 1.} 
While developing a basic chatbot using LLM APIs is relatively straightforward, the complexity increases significantly when the conversation requires intensive external knowledge.
A user's question may contain only partial information, while the rest is in the conversation history, such as containing pronouns referring to an object in the previous question.
Using such a question with unclear pronouns as a query to retrieve knowledge will significantly limit knowledge retrieval accuracy, and answering such unclear questions may increase the risk of hallucination.
Another common case is that when the application is built in a specific knowledge context (e.g., QA based on a GitHub repository), the user may tend to omit some meaningful keywords in their questions, without which the LLMs may misuse its internal knowledge to generate improper answers.
Therefore, such applications require mechanisms to clarify and detail the user questions into queries that contain sufficient information for retrieval knowledge and final answer generation. 

\mypar{Challenge 2.} 
Unlike commercial products~\cite{perplexity, kimi}, building a complete powerful data collection and preprocessing pipeline may be too heavy for tasks relying on private knowledge.
Therefore, a lightweight and configurable retrieval mechanism may be a more desirable solution for developers to utilize their local knowledge from different sources with heterogeneous topics and formats.
Besides, when it comes to running time, accurately answering questions with multi-source knowledge also relies on correctly using a subset of knowledge.
However, there is no existing solution for automatically and efficiently handling the knowledge selection while providing flexibility for taking human intervention.

\mypar{Challenge 3.}
Moreover, as a QA application, the final answer generation is arguably the most critical step.
Retrieved content from multiple knowledge sources may easily consume LLM's effective context window.
To maximize the answer performance, correctly identifying the useful retrieved content and organizing them for question answering are the keys.
But it may be hard to tell how one retrieved knowledge piece from a source is more helpful than the other.
Moreover, for the user experience, providing references is believed to be an effective operation to increase the trust level of the generated answer.
However, it is unclear how to achieve it efficiently with minimum costs.

\mypar{Challenge 4.}
Finally, in addition to correctly answering questions, users may expect the system's response time (i.e., latency) to be as short as possible.
Considering the system's needs for necessary query processing, handling multi-source knowledge retrieval, and answer generation, optimization of execution efficiency requires careful design.

In this paper, we present the design and implementation of an open-source solution, \ours, based on AgentScope~\cite{gao2024agentscope} framework. 
The solution is developed with the following key properties to address the challenges:
\begin{enumerate}
    \item 
    \emph{Context management and query rewrite mechanisms to enhance retrieval accuracy and conversation coherency.}
    \ours provides a built-in \emph{conversation context} query enrichment mechanism that can fill the semantic gaps in the user's current question based on a long conversation history.
    A set of configurable \emph{knowledge context} query rewrite mechanisms is provided for the developers to further adjust the query to fully unleash the power of the retrieval mechanisms and match the most desirable knowledge.       
    
    \item 
    \emph{Efficient routing and retrieval with heterogeneous knowledge sources.}
    \ours supports various knowledge sources with different data topics.
    \ours is also equipped with an efficient routing mechanism that can ensure when a query comes, only the most appropriate knowledge sources will be used to minimize the overall cost while ensuring the answer quality.
    The routing mechanism is also configurable in the sense that developers can provide extra manually written mix-in or scale the score to bias the selection according to their preference and actual use cases.

    \item 
    \emph{Simple yet effective answer and reference generation.}
    We provide a filtering mechanism that is compatible with content from different knowledge sources so that the LLMs can effectively use their window size to generate the final answer.
    Recognizing that providing citations and references is critical for ensuring trust and transparency in knowledge-intensive applications, \ours also implements a straightforward yet reliable strategy for generating citations without requiring model training or introducing additional latency for the central answer generation.

    \item 
    \emph{Configurable and low-latency multi-agent pipeline.}
    \ours employs a multi-agent pipeline architecture. 
    While the pipeline is configurable with three different types of agents, we provide an optimized configuration with parallelization to maximize efficiency and resource utilization.
\end{enumerate}

In summary, this paper introduces \ours, an open-source framework designed to address the challenges of integrating RAG techniques into real-world applications. 
By enabling versatile query enhancement mechanisms with conversation and knowledge context, handling heterogeneous knowledge sources, and ensuring reliable citation generation, \ours seeks to improve the utility and trustworthiness of RAG-based systems. 
Through the practical use cases of \ours, we demonstrate its robustness and adaptability to building effective QA chatbots and other RAG-driven applications, advancing the state of the art in this rapidly evolving domain.
\section{Preliminary}

\subsection{LLM-based Multi-Agent System}
\mypar{Agent.}
In this paper, ``agent'', abbreviated from ``LLM-based agent'', is usually characterized as a paradigm of LLM-centric applications that employ LLM to mimic human brain~\cite{agent_intro}.
With the reasoning capability of LLMs~\cite{wei2022chain, yao2024tree, liu2023llm+, lu2024chameleon}, an agent can decompose a complex task into subtasks that can be solved more reliably;
meanwhile, LLM can make decisions so that an agent can dynamically decide when and how to use tools (i.e., APIs of different external functionalities, such as query portal of current weather of a city) given a task~\cite{yao2022react, parisi2022talm, schick2023toolformer}; finally, with a memory for maintaining context and related knowledge~\cite{shinn2024reflexion, hatalis2023memory}, agents can generate appropriate answers in long conversations or utilize external knowledge.
In this paper, we relax the definition of an agent so that it may contain only a subset of these three elements.

\mypar{Multi-agent and pipeline.}
At the current stage, an agent's performance is not satisfying given complicated tasks.
For example, end-to-end code generation~\cite{hongmetagpt} or have a large number of tools for selection.
Multi-agent systems represent an emerging paradigm, where multiple agents, each specialized in some specific tasks, collaborate to solve complex tasks. 
Some of the multi-agent frameworks are conversational~\cite{wu2023autogen, li2023camel}, where multiple agents collaborate and communicate with each other by generating messages in natural language generated by LLM.
Some other multi-agent frameworks~\cite{hongmetagpt} emphasize the diverse capabilities of LLMs to break down tasks into modular components, allowing individual agents to specialize in specific roles such as information retrieval, reasoning, or decision-making. 
By communicating and exchanging intermediate outputs, agents collectively achieve goals that exceed the capabilities of a single LLM operating in isolation. 
There are also many multi-agent systems that are designed for specific tasks, including medical~\cite{tang2023medagents}, coding~\cite{yang2024swe}, and evaluation~\cite{chan2023chateval}.
Although the above work may have implementation differences, the main idea is still to decompose complex tasks into simple subtasks and let each agent work as an information processing node in a \emph{pipeline}.

\subsection{Knowledge-intensive QA}
RAG is one of the most popular techniques coupled with LLMs, designed to address knowledge-intensive tasks, particularly those requiring high-confidence answers or involving private knowledge unavailable during the model's training phase. 
In the early stages, when language models lacked strong in-context learning capabilities, RAG techniques primarily relied on training or fine-tuning models to integrate retrieved knowledge~\cite{izacard2020leveraging, borgeaud2022improving}. 
However, with the rapid advancement of language models and their demonstrated ability to perform in-context learning, a more efficient and cost-effective approach has emerged. 
This strategy involves appending retrieved text chunks as additional prompts to the LLM input, avoiding the need for task-specific fine-tuning~\cite{ram2023context}. 
In parallel, significant research has been dedicated to improving retrieval mechanisms to ensure higher-quality inputs, such as dense retrieval methods that enhance the relevance of retrieved text~\cite{karpukhin2020dense}. 
These developments highlight the ongoing evolution of RAG techniques and their critical role in extending the capabilities of LLMs for real-world applications.

\section{Knowledge-oriented QA System Designs in \ours}
\ours focuses on the scenario where application users expect to obtain accurate answers based on knowledge from multiple homogeneous or heterogeneous sources.
We use the following hypothetical use case to demonstrate the challenges of building knowledge-oriented QA applications in practice.
The real use cases are presented in the following Section~\ref{sec:usecases}.

\mysubpar{\underline{Hypothetical use case.}}
If a developer of a GitHub repository wants to build  an LLM-based QA plugin based on his repository, he may need to consider the following potential questions:
\begin{enumerate}
    \item Technical questions that LLMs can answer if enough locally available knowledge is provided, such as the code snippet, API documents or tutorials available in this GitHub repository;
    \item Questions related to this GitHub but requiring knowledge beyond the locally-host one, such as whether there are any existing applications built based on this GitHub repository but implemented by a third party, whether there is any third-party solution for some issuing when using this repository, or whether is any third-party comments about this repository.
\end{enumerate}

Inspired by this hypothetical use case, we identify the following problems in building a knowledge-intensive QA system for practical use cases. 
Our system design focuses mainly on solving these problems.

\mypar{(1) Technical questions may not always be formulated clearly.}
There exists a gap between many research-oriented Retrieval-Augmented Generation (RAG) benchmarks and practical scenarios. 
When application users encounter technical issues, they often struggle to articulate their questions accurately, particularly in terms of using the correct terminology, for example, the meaning of an input parameter of a function in the context of the specific GitHub repository in the hypothetical use case above. 
Application users may require additional "warm-up" conversations to clarify and refine their inquiries in these cases. 
The missed information is expected to be filled in by understanding the conversation and knowledge contexts for better retrieval and QA performance.

\mypar{(2) Different questions may prefer different knowledge sources.}
Consider the first type of question in the hypothetical use case above. 
A pipeline built with processed GitHub repository data stored in a local vector database can be an economical and flexible solution that can solve most of the questions.
For the second type of question, collecting information through some online search APIs can provide LLMs with more comprehensive information to generate reliable and up-to-date answers.
In addition, different knowledge sources may also require different query preprocessing techniques and information post-processing operations.
For example, when using online search engines, it may be more important to propose a set of correct keywords based on the complete sentences in natural language, but complete but more organized sentences may be preferred when using embedding similarity-based search in vector databases.

\mypar{(3) Irrelevant information need to be filtered and adopted information need to be cited explicitly.}
The retrieval mechanism cannot always guarantee that all the retrieved content is helpful for answering the question.
The useless information need to be filtered out before being fed into the LLMs as the LLMs have limited effective context windows to process information.
On the other hand, existing LLMs can still generate hallucinated messages or cannot generate perfect answers for some complicated questions in conversations (e.g., code generation with very specific requirements).
As many commercial closed-source solutions (e.g., Perplexity\cite{perplexity}), providing information sources (e.g., links to GitHub files or websites) has been proven as a common way to accommodate such limitations, as the application users can further refer to those provided links for verification or proceeding their difficult task with ground-truth knowledge support.

\mypar{(4) Application users desire similar response latency as simple RAG solutions.}
Another critical metric that influences user experience is response latency.
Application users generally expect LLM applications to deliver responses with consistent speed, regardless of the complexity of the backend information processing. 
However, a significant portion of RAG application developers use only the available LLM APIs and cannot control how model inference optimization works.
This expectation introduces a significant challenge in scenarios where multi-stage information processing is required. 
Despite the inherent complexity of such systems, maintaining efficiency remains a crucial metric that must be considered.

\subsection{Modules and System Structure Overview}
\label{subsec:agents}
\begin{figure}
    \centering
    \includegraphics[width=0.85\linewidth]{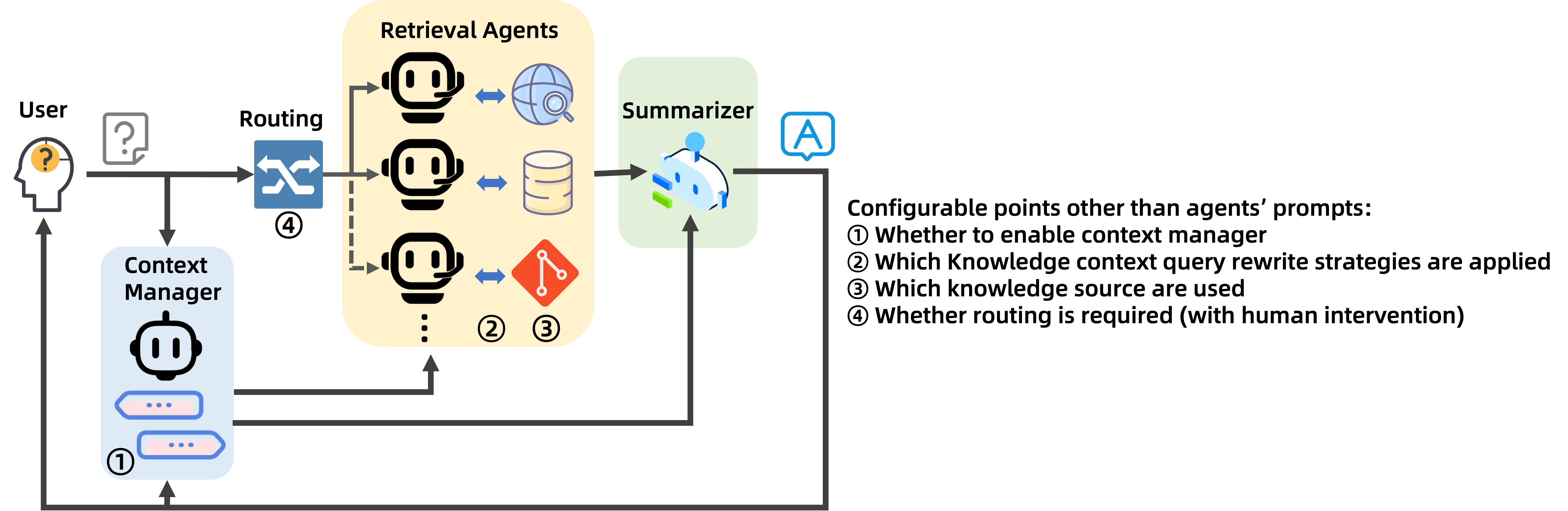}
    \caption{\ours system with agentive modularization and configurable pipeline.}
    \label{fig:overall}
\end{figure}

\mypar{Agentive Modularization.}
In our design, three modules serve as cores in \ours:  conversation context management, information retrieval, and final answer generation.
As in Figure~\ref{fig:overall}, we design a specialized agent class for each module: context manager, retrieval agent, and summarizer.

\begin{itemize}
    \item
    \textit{Context manager}: 
    In everyday communication, conversations often rely heavily on contextual information. 
    The same is true in knowledge-intensive conversational QA tasks.
    In general, the context manager is designed to enrich the query by extracting missing information from the conversation context, ensuring its completeness.
    For instance, when \ours is used to provide QA services for the AgentScope~\cite{gao2024agentscope} GitHub repository, a user might query whether there is any multi-agent games application in the repository. 
    An initial response can be ``Yes. For example, there is a simulation for the game Werewolf that...''
    After that, the user may ask a follow-up question, ``Where can I find the code for it?'' In this case, the context manager must rely on context to determine that ``it'' refers to the Werewolf game in the AgentScope repository and enrich the user question for the following operations.

    \item 
    \textit{Retrieval agents}: 
    Each retrieval agent has its own information source(s) (e.g., similarity search with a local vector database or used by an online search engine). 
    One of the primary tasks of these agents is to further revise the query based on the agent's specific knowledge context and retrieve relative information.
    For example, when the retrieval agent has access to an online search engine, it rewrites the query into a set of keywords to align the search mechanism. 
    In contrast, when a retrieval agent relies primarily on a vector database with dense and sparse retrieval, a sentence enriched with conversation context and task background can improve the matching process during retrieval.

    \item 
    \textit{Summarizer}:
    The summarizer is designed to finalize the answer to a query by integrating information from the context manager and retrieval agents. 
    The goal of the summarizer is to generate content that is both faithful to the information provided by the retrieval agents (i.e., faithful) and appropriate within the context of the conversation (i.e., helpful).
\end{itemize}

\mypar{Configurable end-to-end pipeline.}
While three types of built-in agents are provided and their system prompt can be adjusted according to the application, there is still large room for configuration for different kinds of knowledge-intensive QA applications.
The following points are exposed for configuration by developers.

\begin{itemize}
    \item \textit{Whether to enable context manager.}
    While the context manager can enrich the user query and provide necessary context information for the final answer generation, analyzing and processing the context will take extra time for LLM.
    In some cases, where extremely low latency is required, the context manager can be disabled to further reduce latency.
    Instead, one can directly provide the entire conversation history for the summary and ask it to generate the final answer directly.
    Of course, if the context manager is disabled, the quality of the final answer may be hard to control if the conversation is complicated or the LLM is weaker.

    \item \textit{Which knowledge context query rewrite strategies are applied.}
    As discussed above, query rewrite can be a critical point for a knowledge-intensive application, helping to retrieve more accurate content from knowledge sources.
    \ours provides interfaces in the configuration with multiple build-in rewrite strategies (and rewrite prompts of some strategies) for the developers.
    More details will be discussed in Section~\ref{subsec:query-rewrite}.
    
    \item \textit{Which knowledge sources are used.}
    We provide built-in vector database and online search engine APIs as the two built-in knowledge sources for the application.
    For locally hosted knowledge, \ours inherits LlamaIndex~\cite{llamaindex} functions and is integrated into our system so that developers can switch different functions by just changing their knowledge configuration files.
    Details are discussed in Section~\ref{subsec:routing+retrieval}.
    
    \item \textit{Whether routing is required (with human intervention).}
    We also prove some flexibility for the routing mechanism.
    Besides the built-in mechanism routing, we allow developers to cast their human preferences for the routing by providing additional manually written mix-in text or scaling the importance of some weights.   
    More details are deferred to Section~\ref{subsec:routing+retrieval}.

\end{itemize}

With these key agents and configurable pipeline design working together, it becomes \ours.

\subsection{Context-based Query Enhancing}
\label{subsec:query-rewrite}

Knowledge-intensive QA applications usually have two kinds of ``contexts'' as essential factors to answer a question: the context of conversation and the context of knowledge sources.
Correspondingly, two types of agents in our design can rewrite the user query to enrich its semantic meaning with different information sources: context manager and retrieval agent.
In general, the context manager, designed to help the retrieval agent and summarize digesting conversation in advance, ensures that the query contains necessary information from the conversation context;
on the other hand, a retrieval agent is supposed to rewrite the query better to fit the retrieval mechanism in its knowledge source context.
While the rewrite mechanisms of context manager are fixed, the ones for retrieval agents are designed to be more flexible.
The following are details about context-based query enhancement.

\begin{figure}
\centering
    \begin{subfigure}[b]{0.48\textwidth}
        \centering
        \includegraphics[width=\textwidth]{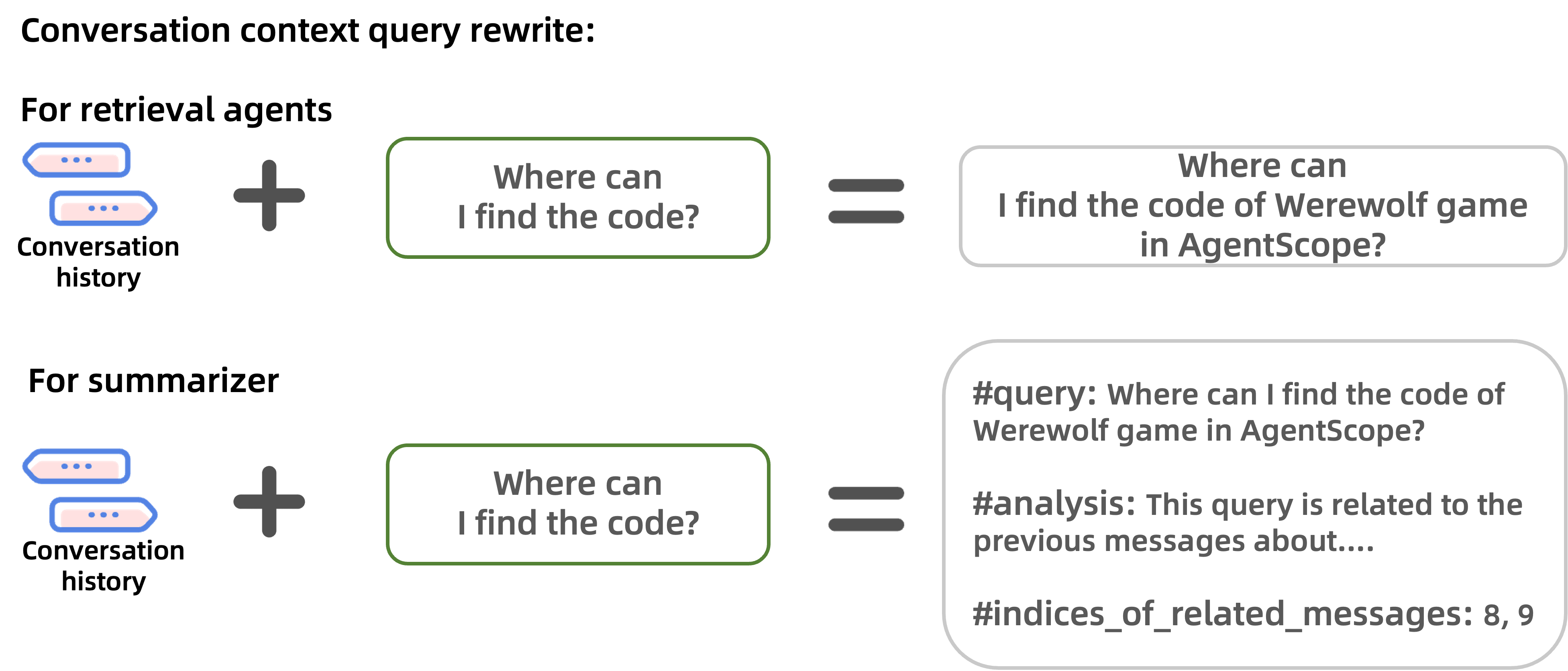}
        \caption{Conversation context rewrite mechanisms}
        \label{fig:conversation_rewrite}
    \end{subfigure}
    \hfill
    \begin{subfigure}[b]{0.48\textwidth}
        \centering
        \includegraphics[width=\textwidth]{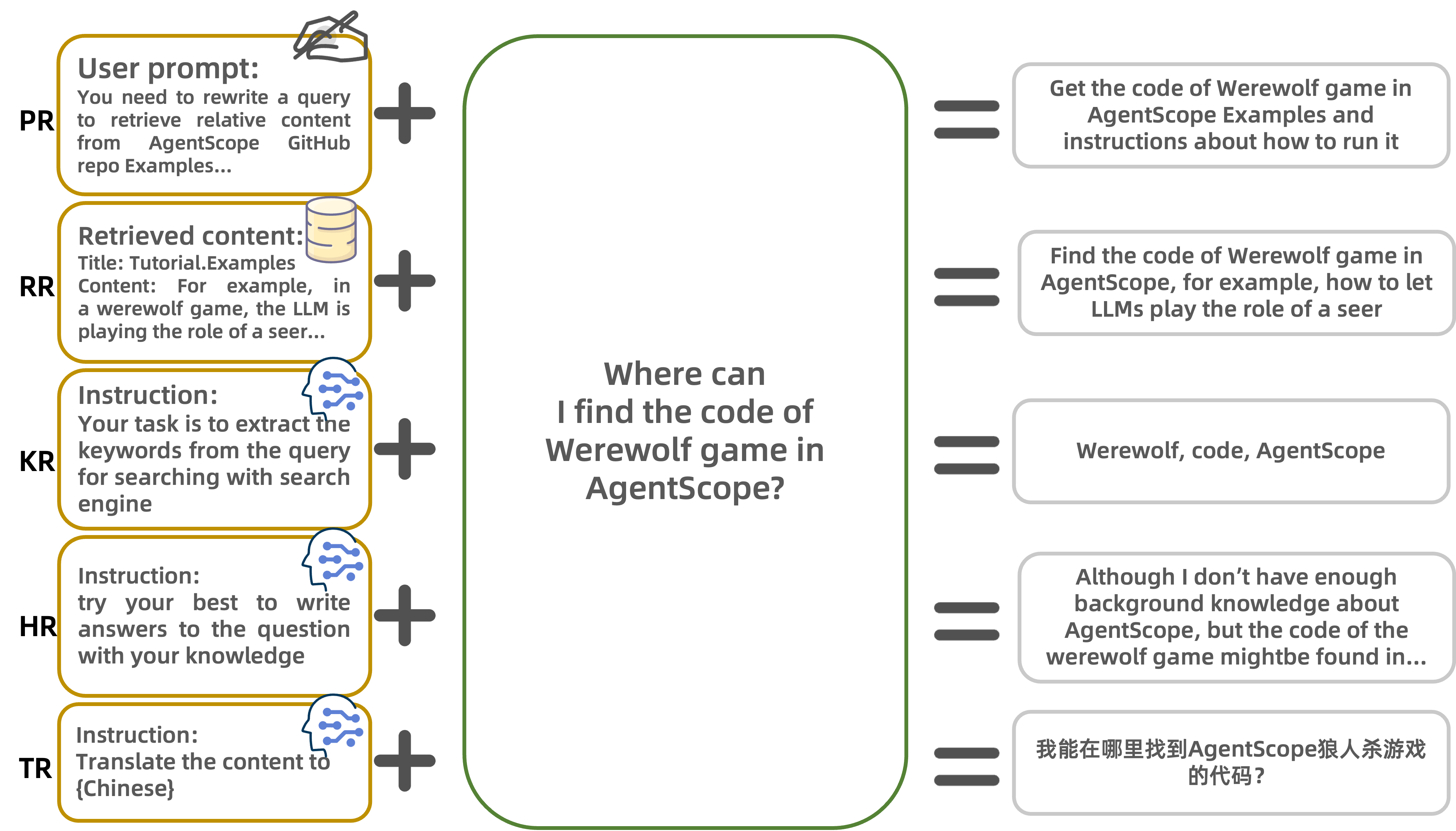}
        \caption{Knowledge context query rewrite mechanisms}
        \label{fig:retrieval_rewrite}
    \end{subfigure}
\caption{Query rewrite mechanisms in \ours.}
\label{fig:rewrite}
\end{figure}

\paragraph{Context manager: conversation-contextual query rewrite for retrieval agents.}
As discussed in Section~\ref{subsec:agents}, understanding the user's real intention behind a query heavily depends on the context of the conversation.
Without some keywords in the conversation context, knowledge retrieval can be pointless.
Thus, the goal of the query rewrite mechanism at this stage is to enrich the query with precise conversation context information.
1) The first point to consider is that some key information is lost in the query but can be obtained from the conversation history (e.g., the example mentioned in Section~\ref{subsec:agents}).
The context manager checks whether the query itself is ambiguous, including containing pronouns referring to terms appearing in the previous conversation.
2) The second consideration is that some users may rephrase their previous questions when they find that the answer provided by \ours is not satisfactory.
Therefore, the context manager also needs to revise the query with reflection.
3) Besides, this step is expected not to consume too much time as one of the intermediate steps in the pipeline.
Therefore, these two goals are integrated into the LLM prompt together with the conversation history and the LLM is expected to properly rewrite the question with necessary details filled and emphasized.
Considering the time constraint, we also recommend using a lightweight LLM to handle this task.

\paragraph{Context manager: conversation-contextual query rewrite for summarizer.}
The final step in generating the answer is another step that is heavily based on the context information.
In this step, the quality of the generated answer depends directly on the degree to which the LLM understands the relevant context.
Our observations indicate that while many LLMs perform well for requests with simple requirements, but they may struggle to handle more complex requests involving multiple sub-tasks (e.g., understanding the conversation and answering the current question with given context in one generation).
Besides, when the context is long (for example, a long answer is provided to the previous query), including the entire conversation history in the final step can unnecessarily consume the LLM's effective context length, reducing efficiency and potentially affecting output quality.
Nevertheless, the final answer generation requires more detailed information to generate the final answer, which may not always aligned with the rewrite goal for retrieval agents.
To address these challenges, a second function of the context manager is to perform a detailed and reliable analysis of the conversation context, and only the digested conversation context will be passed to the summarizer for final answer generation. 
This ensures that only the most relevant information is retained, improving the overall quality and efficiency of the final response generation.

More precisely, we prompt the context manager to generate answers with two fields: \texttt{analysis} and \texttt{indices\_of\_related\_messages}.
Generating \texttt{analysis} for the relation between the history and the current query can be considered with a similar effect to the Chain of Thoughts~\cite{wei2022cot}, which enables LLMs to generate answers more rationally.
After the \texttt{analysis}, LLMs should have higher confidence about which previous historical conversation piece is indeed related to the current one.
To provide more precise information for summarization, we also prompt the LLM to generate \texttt{indices\_of\_related\_messages}, with which we can directly extract the related messages by the indices.

\paragraph{Retrieval agent: knowledge-contextual query rewrite.}
In \ours, different retrieval agents have different knowledge sources with potentially heterogeneous data types.
In order to maximize the accuracy of information retrieval, different query rewrite prompts or even query rewrite algorithms can be employed at the agent level.

The built-in rewriting strategies are listed as follows and illustrated in Figure~\ref{fig:retrieval_rewrite}.
\begin{itemize}
    \item \texttt{Prompt rewrite (PR):} Rewrite the query following the instructions of the prompt.
    This mechanism allows developers to customize the prompt to LLM inject necessary information to the query, typically prior knowledge or understanding of the knowledge.
    
    \item \texttt{Retrieval rewrite (RR):} Rewrite the query with some knowledge content retrieved with the original query.
    An intuition of this method is that LLMs can help revise more accurate queries when providing some knowledge context.
    
    \item \texttt{Keyword rewrite (KR):} Extract only keywords from the original query for search engines.
    When using online search engine APIs, keywords are usually the best way to perform a search, as the redundant information in the original query may diverge the search to unrelated content.

    \item \texttt{HyDE rewrite~\cite{gao2023hyde} (HR):} Generate a paragraph with LLM's internal knowledge to answer the query, then take the paragraph as the new query.
    It is shown that in some cases, the embeddings of the LLM generated response without the interference of external knowledge can be more similar than the one of the raw query to the desired embeddings of the desired knowledge, so that it can improve the retrieval performance.

    \item \texttt{Translation rewrite (TR):} 
    Rewrite the query in the same language as the one specified in the configuration of the knowledge source.
    For the cases where the language in the knowledge source is different from the one of the query, it is believed that maybe mapping it first to the targeted language can provide a more accurate retrieval result.
\end{itemize}

All the above rewrite strategies are available and only require changing configuration files. 
However, developers can provide their own rewrite strategies and easily integrate them into their applications.

\subsection{Efficient Multi-source Information Retrieval}
\label{subsec:routing+retrieval}

\subsubsection{Knowledge retrieval mechanisms}
The retrieval agents play key roles in \ours because they have access to knowledge sources.
Their responsibility is to retrieve and provide relative knowledge to user queries.
The following built-in support for the different knowledge sources can be easily assigned to retrieval agents via configuration.
\begin{itemize}
    \item \textit{Local knowledge stored in vector databases (VDB).}
    Using local vector databases is the most popular and efficient method to construct local knowledge bases, especially after LLMs demonstrate their in-context learning capability.
    The embeddings of the knowledge in VDB only need to be computed once and used to construct the indexing.
    At inference time, it only requires the embedding model to encode a query into a vector for a similarity match.
    The storage of local knowledge and retrieval from local VDB has a few advantages.
    The first advantage is that it can utilize local knowledge, which is available locally, with little data privacy or data sovereignty concerns.
    A second advantage is that VDBs usually support a mixture of dense (embedding similarity) and sparse (BM25) retrieval, which can provide superior performance when semantic similarity is the only metric for retrieval.
    A third advantage is that even if the embedding relies on API, generating embedding for a query is usually much more affordable (even if it can be done locally) than calling the API of search engines.
    In \ours, users can choose to use LlamaIndex~\cite{llamaindex} built-in in-memory VDB or  Elasticsearch~\cite{elasticsearch2018elasticsearch}.

    \item \textit{Online search engines.}
    There are many advantages of using online search engines as an information source. 
    One is that it can guarantee the information is up-to-date.
    A second advantage is that the search engine algorithms rank the returned results, which consider many different factors (e.g., timeliness, popularity and authority with Internet-scale information).
    Therefore, even if it is more expensive than the local VDB method, online search engines are still one of the popular knowledge sources in many applications.
    The built-in support in \ours for search engine is built on Bing search~\cite{bing}, but developers can easily switch to Google search and others.

    \item \textit{Domain specific HTTP API.}
    Some websites provide in-site search APIs, which can return some domain-specific knowledge presented on the website but are not open to web crawlers.
    To leverage such APIs, we provide some flexible request-response parsing functions that facilitate the calling step and response parsing steps.
\end{itemize}

\paragraph{Discussion: raw text information or LLM digested information?}
Determining whether the retrieval agent should process the retrieved information before feeding it into the summarizer is tricky.
The answer should be given case by case, depending on many different factors.
For example: 1) Can the LLM used by the summarizer support processing long context information and still provide reasonably good reasoning capability?
2) Do we expect that the final answer of \ours can be generated efficiently (e.g., seeing the first token within ten seconds)?
If the answers to both questions are ``yes'', then letting the retrieval agents return the raw text and letting the summarizer process the raw information directly may be more appropriate because it can avoid additional latency because of LLM processing.
Besides, such a more straightforward approach can reduce the chance of introducing LLM hallucinations.
However, suppose only context-length-limited LLMs are available, or the available LLMs cannot provide acceptable performance in long-context reasoning; it may be a better solution to let each retrieval agent process the raw retrieved information, extract the key information first, and only pass the valuable information to the final answer generation step.

\subsubsection{Embedding clustering routing}
We expect each retrieval agent to take charge of one or a few knowledge sources.
The key guidance is that one can group similar knowledge sources to the same retrieval agent so that the knowledge context query rewrite can benefit the retrieval of all similar knowledge sources.
On the other hand, knowledge sources with very different topics or contents are supposed to be assigned to different retrieval agents.

However, a challenge is correctly selecting the best-fit retrieval agent(s) with the most relevant knowledge source.
Most existing multi-agent routing mechanisms rely on 1) manually created descriptions for the (functionalities of) agents and 2) using LLMs as decision-makers to decide which agents should be activated to provide knowledge.
However, such a combination is not suitable for routing between knowledge retrieval agents.
As a knowledge source may contain a large volume of knowledge, it can easily exceed the context length of any LLMs; nevertheless, it is challenging for human beings to summarize all the knowledge from a source comprehensively.
Meanwhile, using LLMs to select the knowledge source may not be a good idea in practice because 1) its output can be non-negligible randomness, which can make it hard to ensure consistency; 2) LLMs inference can be slow and restricted by the context length (i.e., not suitable for high-efficiency applications or context with long conversation).

\begin{figure}
\centering
\includegraphics[width=0.6\linewidth]{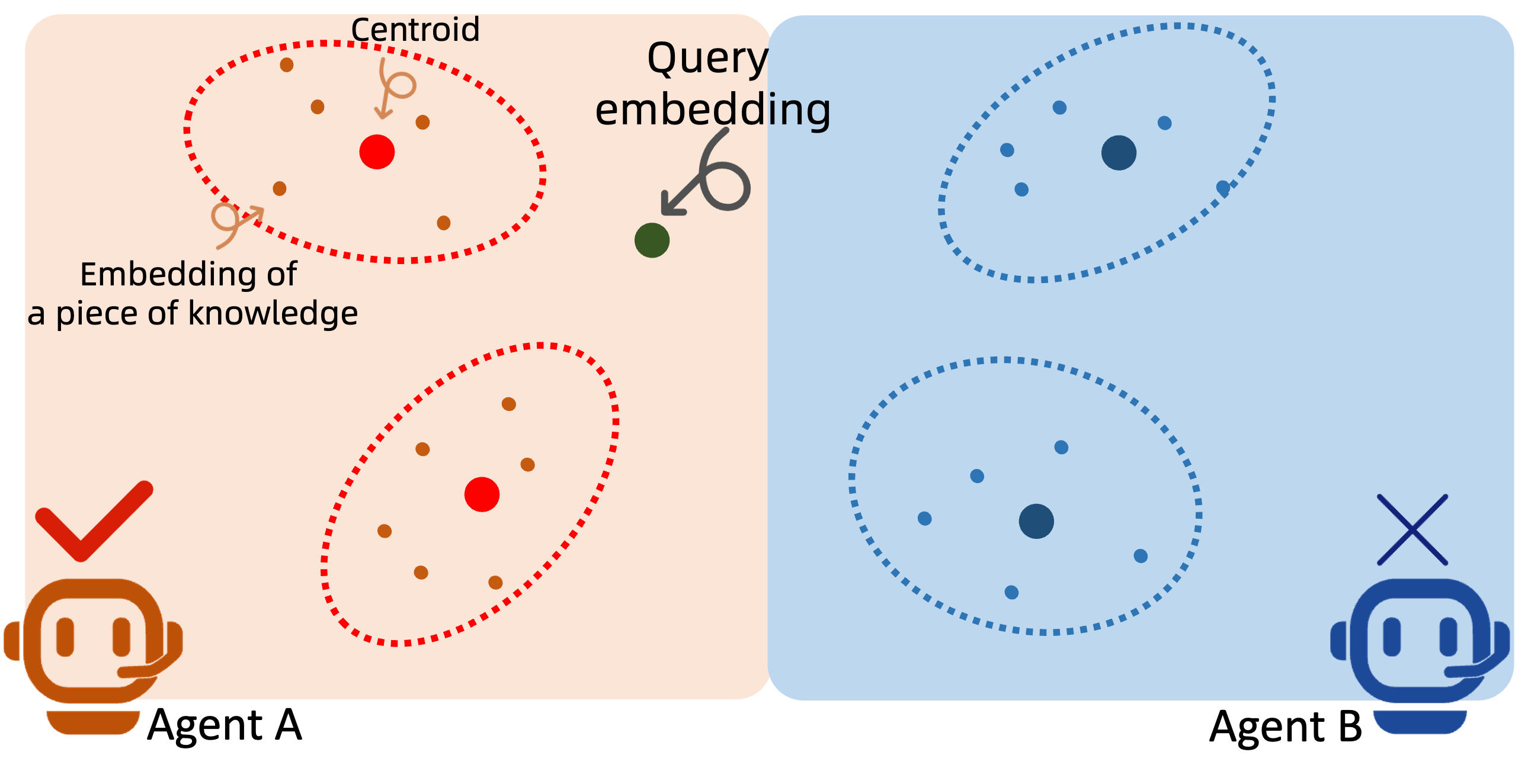} 
\caption{A simple visualization of the routing mechanism. Because the query embedding is closer to centroids in Agent A's knowledge domain, Agent A is roused to conduct knowledge retrieval.}
\label{fig:routing}
\end{figure}

\paragraph{Backbone routing mechanism.}
We adopt an algorithm similar to \citep{feijie}.
Figure~\ref{fig:routing} gives a simple visualization of the core idea.
The key idea is to utilize the embeddings of the knowledge in each agent.
In addition to being used in retrieval, the embeddings of the knowledge (e.g., text chunks) are also perfect representations of that knowledge.
Therefore, the embeddings of chunks of knowledge are first clustered in each agent, and the centroids of the cluster are considered to be \emph{synopses} of the knowledge.
When routing, an embedding of the query needs to be used for similarity search with the centroids of all retrieval agents. 
Only the agents with top-$K$ similar centroids to the query embedding are activated for exact knowledge retrieval.

\paragraph{Manual mix-in.}
Different from \citep{feijie}, we also need to consider retrieval agents using online search APIs as the knowledge source, which has limited or even no local knowledge to guide the selection of knowledge.
In addition, the method of in \citep{feijie} is completely data-driven.
While complete data-driven is a desired feature, it also means difficult to impose human preferences when it comes to real practice.

To resolve such inconveniences, we allow developers to either 1) provide the initial description or 2) provide a set of typical knowledge chunks (e.g., QA pairs or chunks of plain text) as mix-in to enrich the routing.
Such manual descriptions are also encoded into embeddings. 
When routing, the similarity considered becomes a weighted average of the similarity score of the local knowledge pieces to the query and one of the manual mix-ins to the query.
When developers have higher confidence in their manual description and want it to dominate the process of the selection, a higher weight can be assigned to the similarity score of the manual description; if the developer wants to save effort and have plenty of coherent knowledge in each agent's knowledge base, a higher weight can be assigned to the similarity score of the local knowledge pieces.
If the knowledge source is not local, the manual description can serve as the only centroid of the knowledge.

\paragraph{Score scaling.}
In practice, the similarity scores when comparing the same query with different types of documents can vary in various ranges.
For example, when matching a natural language query with Python code knowledge, it usually has lower scores than matching with natural language documents, even if the key relative information related to the query is indeed in the Python code.
To handle such cases, we employ a simple but effective strategy that allows users to scale some similarity score related to specific types of knowledge, either up or down.

\subsection{Summarization}

\paragraph{Reranking for filtering.}
A common strategy to avoid the false-negative cases (i.e., high-relevance knowledge is not retrieved) is to retrieve slightly overwhelming information that is larger than the context length of the LLMs.
In \ours, since multiple retrieval agents can provide different knowledge sources, the summarizer can receive overwhelming information that cannot fit into the effective context window of LLMs.
However, although embedding similarity matching mechanisms can efficiently retrieve a lot of relevant information, they are not metrics to rank or filter information due to the following important issues.
1) Some retrieval agents may even never compute the similarity score, such as those using the online search engine as the knowledge source.
2) High similarity scores can contain false positive signals as they are computed in compressed vector space.
3) The similarity scores from different resources may not be comparable because the retrieval agents may rewrite the query for some purpose, so the similarity scores used in retrieval are actually based on different queries.
4) The nature of knowledge can affect the outcome, such as the data format, chunk size, etc.

Reciprocal rank fusion (RRF)~\cite{cormack2009reciprocal} can be a model-free statistic that may help in reranking and filtering. 
However, it is unlikely to have duplicated knowledge pieces retrieved from different knowledge sources.
Therefore, RRF may not be that useful for reranking in multi-source knowledge retrieval.
Instead, we use the reranking model to sort the raw retrieved fragments.
Although a reranking model can introduce additional computation cost, it is a more reliable and general method for our tasks.

\paragraph{Citation generation.}
Citations and references can provide additional confidence for the generated content as users can verify the answer by looking at the references provided.
However, generating citations is a challenging task for LLMs \cite{gao2023enabling}.
In \ours, our goal is to generate citations efficiently without training a specific model or introducing a significant regression in latency or reasoning performance for the final answer.

We tested several approaches. 
Our initial is \emph{one-step}, with which the LLMs are prompted to finish the following tasks in a single answer: 1) analyze which chunks can help answer the query question; 2) generate the answer to the query question; 3) select the indices of the related chunks.
Then, the final answer is to assemble the generated answer and extract the reference with the indices.
However, such an approach has several limitations.
First, the LLM must have strong reasoning and context-processing ability.
Second, the LLM must generate structure output (e.g. in JSON format) from which different information can be extracted, which means that it is impossible to use the stream mode of the LLMs to provide a low-latency experience for users and introduce additional task failure risk as LLMs cannot always correctly format its output.
Third, as the retrieved knowledge is usually in chunks, they can be from different or the same information source (e.g., the same paper).
Therefore, it requires additional effort to handle or distinguish duplications.

\textit{Solution: A look-back approach.}
To bypass the above problems, we design a look-back strategy for citation generation.
The entire generation utilizes the stream mode provided by many LLM APIs or local inference frameworks and consists of two stages.
The first stage is to prompt the LLM to generate an answer to the query question based on the retrieved knowledge.
The answer is presented directly to the users.
The second stage is to provide the generated answer together with the retrieved knowledge to the LLM and let it output the reference of the knowledge used to generate answers.
Such a strategy is robust with weaker requirements on the reasoning capability of LLMs.
Besides, as the content is generated in stream mode, the user's experience can be significantly improved as the first token they observed at the same time it is generated, and only a small pause will be observed after the answer is generated and waiting for the citation generation.

\subsection{Optimizing Pipeline for Efficient Execution}
Many LLM-based applications seek low response latency, which poses a challenge for multi-stage applications like \ours.
More specifically, since multiple sub-tasks exist, LLM will inevitably be used to generate multiple times and introduce some intermediate result waiting time. 
Therefore, efficiently arranging the execution requires parallelizing as much LLM usage as possible.
In order to optimize efficiency, the execution of \ours can be executed in parallel with requests for asynchronous model APIs as follows.
\begin{itemize}
    \item \textbf{Parallelization 1: Query ingest.}
    At this stage, two functions are executed in parallel.
    One is the \emph{context manager conversation context query rewrite for retrieval agents}, which fills in missing key information for the current query question based on the whole conversation history so that the following retrieval stage can use more accurate information.
    The other is \emph{query routing}, deciding which retrieval agents are the correct ones to execute.

    \item \textbf{Parallelization 2: Knowledge retrieval and context analysis.}
    At this stage, each retrieval agent obtains the query enriched with the necessary context information of the conversation.
    At this stage, each agent can rewrite a knowledge-context query to match the knowledge context and retrieve knowledge.
    Meanwhile, the context manager analyzes the context of the previous conversation and generates distilled information to ensure the final answer fits the conversation.
    All of the above agent operations are performed in parallel.
    At the end of this stage, the retrieval information and the conversation context analysis results are shared with the summarizer for the final response generation.
\end{itemize}

\section{Use Cases}
\label{sec:usecases}

\begin{table}
\centering
\begin{tabular}{|p{2cm}|p{2cm}|p{2cm}|p{3.5cm}|p{3.5cm}|}
\hline
\textbf{Use case} & With context manager & Routing / Preference adjust & Offline knowledge source(s) & Online knowledge source(s) \\ \hline
AgentScope QA   & Yes    & ON / No    &AgentScope tutorial, code, examples, API docs and FQA set & -      \\ \hline
ModelScope QA     & Yes      & ON / Yes   & ModelScope tutorials, and 5 GitHub repos  &ModelScope offical article, models, datasets,   \\ \hline
Olympic Bot    & No      & OFF/-     & - & Olympic events      \\ \hline
\end{tabular}
\caption{Use Case Configurations}
\label{tab:use_cases}
\end{table}

\mypar{Sytem implementation.}
\ours is implemented based on a multi-agent framework, AgentScope~\cite{gao2024agentscope}.
At the agent level, the agents inherits from the built-in agents in AgentScope but with extra features specialized for \ours, such as the methods supporting the efficient routing mechanism and some external knowledge management and retrieval functions.
At the pipeline level, the agents receive input and pass process information via AgentScope's message objects; some of the parallel execution stages in the pipeline are tailored specifically in \ours for easy management and code management. 

In the following, we demonstrate how we configure \ours to build different QA applications for three different tasks.

\subsection{Small scale application: AgentScope QA}\par

As a proof-of-concept example, we first present an example with offline knowledge sources with different data formats and topics as a starting point to demonstrate how \ours works.

\begin{wrapfigure}{r}{0.25\textwidth}
  \vspace{-0.95cm}
  \begin{center}
    \includegraphics[width=0.24\textwidth]{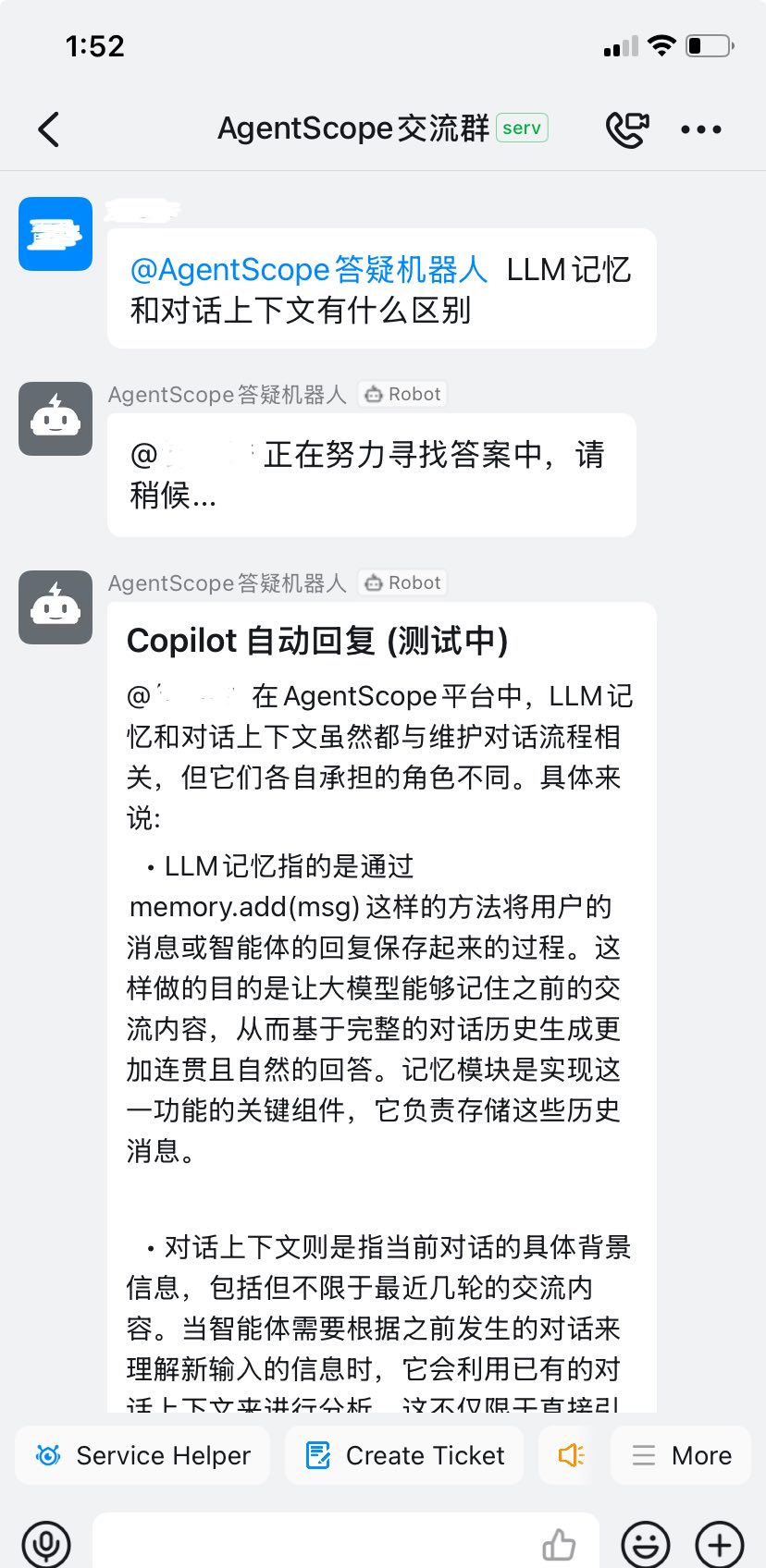}
  \end{center}
  \vspace{-0.2cm}
  \caption{Use case in Q\&A group of AgentScope}
  \label{fig:as_use_case}
  \vspace{-1.5cm}
\end{wrapfigure}

\mypar{Goals.}
In this use case, we adapt \ours to help answer questions about AgentScope's GitHub repository.
The expectation for this application is to serve as a chatbot in a Q\&A group for developers building their multi-agent application with AgentScope framework, providing an accurate response in time to the raised questions.
It is observed that the most common questions fall in the following categories:
\begin{itemize}
    \item \textit{Preliminary project questions.}
    As the Q\&A group is open for everyone, some potential users of AgentScope are also presented in the group. 
    Their questions are usually about the feature of AgentScope, its advantages compared with other similar open-source frameworks, and the feasibility of using AgentScope for their task.
    Once some of them decide to initiate their project with AgentScope, they may be looking for whether there are already demonstration examples for tasks similar to theirs provided in the GitHub repository for reference.

    \item \textit{Coding or debugging related questions.}
    Another majority of questions in the Q\&A group appear to be developers using AgentScope but encounter issues in their development process.
    For example, they may seek clarifications of two different agent classes or help to solve the execution errors in their AgentScope-based applications.
\end{itemize}

\mypar{Application Configuration.}
To satisfy the above goals and provide reliable answers to the questions in Q\&A group, we configure our \ours in terms of knowledge and execution as the following.

\begin{itemize}
    \item \textit{Knowledge sources configuration.}
    The knowledge bases equipped to the agents in this use case span the content related to the core functions of library: \underline{the tutorial} (Markdown files), \underline{code} (mainly Python code), \underline{API documents} (processed into text files).
    The \underline{examples} in the AgentScope library, which serve as good demonstration examples for beginners, are also included as knowledge of how to use the AgentScope framework with code and description in Markdown format.
    Besides, some of the frequently asked questions beyond the scope of the repository (e.g., comparison with other frameworks) are gathered manually, and are summarized and filled with appropriate answers to form a \underline{FQA set} as an additional knowledge source to the repository. 
    All of this knowledge is processed (i.e., chunking and generating embeddings) and stored in a vector database for query by a knowledge configuration file.
    Each type of knowledge is assigned to a retrieval agent.

    With the routing mechanism, the preliminary project questions are usually answered with the knowledge shared by the agents charging tutorial, examples, and FQA set, while answering the coding or debugging questions will depend on information from tutorial, code, and examples.

    \item \textit{Pipeline.}
    This application is configured with all three types of agents activated: context manager, retrieval agent, and summarizer.
    The context manager is used to help understand the real intention of the user in a conversation context. 
    The retrieval agents are configured with the \texttt{Prompt rewrite} module with prompts design for each agent according to their equipped knowledge.
\end{itemize}

Figure~\ref{fig:as_use_case} is a screenshot of a real QA with \ours AgentScope QA application in the DingTalk Q\&A group.

\subsection{Larger scale: ModelScope QA}
\mypar{Goal.}
While serving as a Q\&A chatbot similar to the AgentScope use case, the spectrum of questions to be handled is significantly larger than the ones in Agentscope.
Modelscope community is a platform of open-sourced machine learning models, datasets, training/finetuning libraries and applications built with LLMs and other models.
It is expected a chatbot to provide an accurate initial answers potentially based on all these kinds of the knowledges from different sources.

\mypar{Application Configuration.}
To helpfully serve the users in the community, we highlight the specialties in configure as the following.
\begin{itemize}
    \item \textit{Knowledge sources configuration.}
    The knowledge sources used in these applications can be categorized into two types, online and offline.
    
    The online knowledge sources include \underline{model} and \underline{dataset} information, \underline{official articles} about the latest open-source community technical progress.
    These knowledge sources are achieved by Bing search API with different constraints, i.e., restricted to domain of available models/datasets/articles on the website.
    It is configured to use a commercial search engine instead of in-site search because the results of Bing can also be used to provide related knowledge that ranked by more sophisticated mechanisms beyond text similarity, such as recommending the most popular Text-to-Image models.

    The offline knowledge sources include the \underline{tutorial} documents and \underline{eight different GitHub repositories} affiliated with ModelScope.
    The tutorial covers knowledge about how to use models, datasets and computation resources available on \url{modelscope.cn}, and the repositories contain code files and repository-level instructions.
    Compared with the online knowledge sources, these knowledge sources can be hosted and retrieved locally because the retrieval standard is more 

    \item \textit{Pipeline configuration.}
    Generally speaking, the pipeline configuration is similar to the AgentScope use case, involving all three kinds of agents.
    But key difference is in the routine mechanism configuration. 
    We provide some manual description mix-in for the routing mechanism to bias some selections manually.
    For example, both the tutorial and model knowledge source contains the information of some models (or models with similar names); however, the information from model knowledge source is more up-to-date than the one from the tutorial. 
    Therefore, we add some manual mix-in to bias the "model recommendation" type question to use model knowledge source rather than tutorial.
    Because some knowledge sources are considered more reliable and official (i.e., official articles and tutorial), we set a scaling factor greater than 1 to make the information pieces from those sources have a higher chance of being selected.
    
\end{itemize}

\begin{figure}
    \centering
    \begin{subfigure}[b]{0.32\textwidth}
        \centering
        \includegraphics[width=\textwidth]{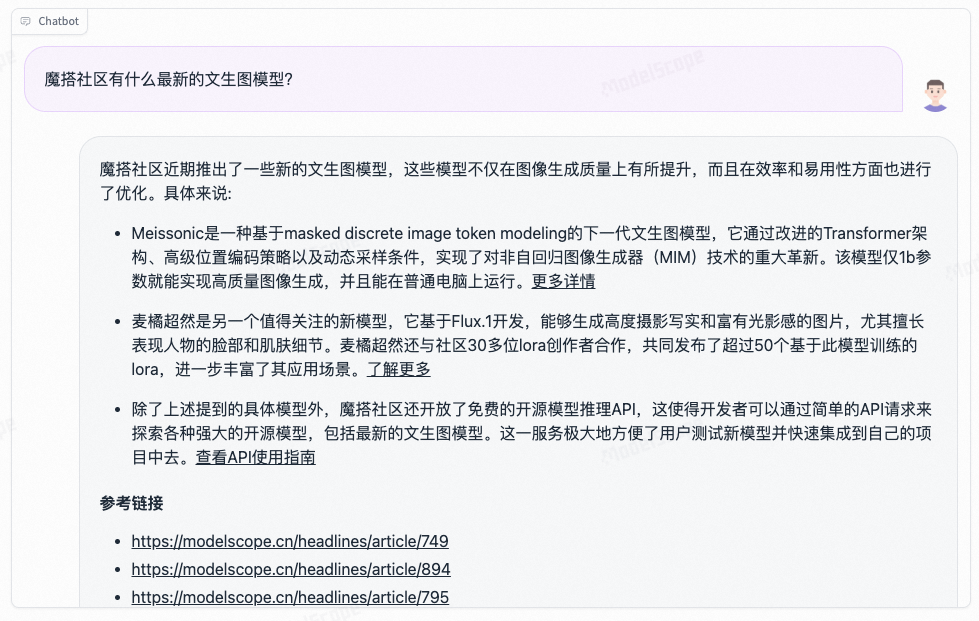}
        \caption{Online sources usage}
        \label{fig:ms_online}
    \end{subfigure}
    \hfill
    \begin{subfigure}[b]{0.32\textwidth}
        \centering
        \includegraphics[width=\textwidth]{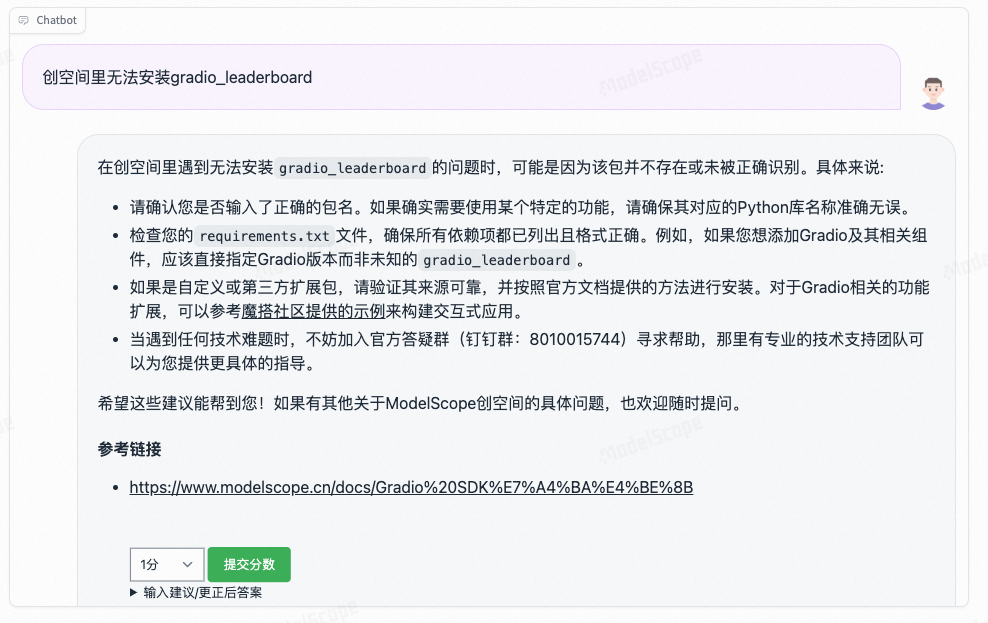}
        \caption{Offline sources usage}
        \label{fig:ms_offline}
    \end{subfigure}
    \hfill
    \begin{subfigure}[b]{0.32\textwidth}
        \centering
        \includegraphics[width=\textwidth]{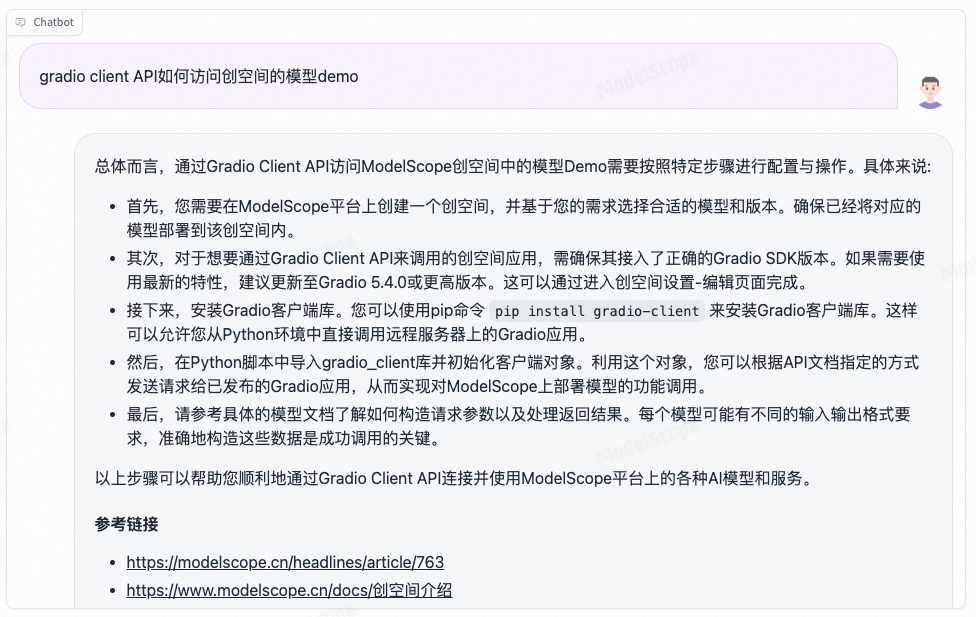}
        \caption{Mixture usage}
        \label{fig:ms_mix}
    \end{subfigure}
    \caption{Three demonstration QA pairs using different knowledge resources in \href{https://www.modelscope.cn/studios/AI-ModelScope/modelscope_copilot_beta}{Modelscope QA}.}
    \label{fig:ms_use_csae}
\end{figure}

Figure~\ref{fig:ms_use_csae} shows three QA pairs from the ModelScope QA using different knowledge sources.

\subsection{Turbo Scale: Olympic Bot on Weibo}

\mypar{Goal.}
During the Paris 2024 Olympic Games, \ours serves as the core of the back-end algorithm to generate auto-comments for the posts related to the Olympics\footnote{\url{https://weibo.com/u/7929611818}}.
The posts and comments of this Weibo bot are supposed to focus only on Olympic Games, including the news and historic Olympic events.
However, there are many other bots performing on Weibo, but there is a restriction that no more than two bots can reply to the same Weibo post.
Therefore, there is a race condition and the response generation efficiency becomes a key point in this scenario.
To encounter such challenge, our configuration of \ours needs to make the following changes.

\mypar{Application Configuration.}
In order to fulfill the requirements, \ours is configured as the following.
\begin{itemize}
    \item \textit{Knowledge sources configuration.}
    We configure a retrieval agent to use the specialized \underline{search API} for Olympic related events.
    The APIs perform searches similar to public search engines using keywords to match related information.

    \item \textit{Pipeline configuration.}
    To extremely reduce response latency, the pipeline configuration becomes simple. 
    The context manager is deactivated, and the only retrieval agent will perform a keyword query rewrite only.
    The full conversation history and the retrieved knowledge will be directly fed to the summarizer to generate the final answer.
\end{itemize}

With such knowledge and pipeline configuration, the end-to-end latency is reduced to less than 10 seconds per post or comment, while the responses are still very informative and popular.

\section{Conclusion}
In this technical report, we introduce \ours, our configurable knowledge-integrated multi-agent system for developers to build their knowledge-intensive QA system.
We present three use cases built on our system with different emphases to demonstrate that \ours can be configured to various applications.
While the current version of \ours focuses on knowledge-intensive QA tasks, future development can expand its capabilities to address a broader range of challenges, including code generation based on some specific local code base and interactive recommendation system for e-business.

\bibliographystyle{plain}
\bibliography{references}

\begin{thebibliography}{10}

\bibitem{borgeaud2022improving}
Sebastian Borgeaud, Arthur Mensch, Jordan Hoffmann, Trevor Cai, Eliza Rutherford, Katie Millican, George~Bm Van Den~Driessche, Jean-Baptiste Lespiau, Bogdan Damoc, Aidan Clark, et~al.
\newblock Improving language models by retrieving from trillions of tokens.
\newblock In {\em International conference on machine learning}, pages 2206--2240. PMLR, 2022.

\bibitem{chan2023chateval}
Chi-Min Chan, Weize Chen, Yusheng Su, Jianxuan Yu, Wei Xue, Shanghang Zhang, Jie Fu, and Zhiyuan Liu.
\newblock Chateval: Towards better llm-based evaluators through multi-agent debate.
\newblock {\em arXiv preprint arXiv:2308.07201}, 2023.

\bibitem{cormack2009reciprocal}
Gordon~V Cormack, Charles~LA Clarke, and Stefan Buettcher.
\newblock Reciprocal rank fusion outperforms condorcet and individual rank learning methods.
\newblock In {\em Proceedings of the 32nd international ACM SIGIR conference on Research and development in information retrieval}, pages 758--759, 2009.

\bibitem{elasticsearch2018elasticsearch}
BV~Elasticsearch.
\newblock Elasticsearch.
\newblock {\em software], version}, 6(1), 2018.

\bibitem{gao2024agentscope}
Dawei Gao, Zitao Li, Xuchen Pan, Weirui Kuang, Zhijian Ma, Bingchen Qian, Fei Wei, Wenhao Zhang, Yuexiang Xie, Daoyuan Chen, et~al.
\newblock Agentscope: A flexible yet robust multi-agent platform.
\newblock {\em arXiv preprint arXiv:2402.14034}, 2024.

\bibitem{gao2023hyde}
Luyu Gao, Xueguang Ma, Jimmy Lin, and Jamie Callan.
\newblock Precise zero-shot dense retrieval without relevance labels.
\newblock In {\em Proceedings of the 61st Annual Meeting of the Association for Computational Linguistics (Volume 1: Long Papers)}, pages 1762--1777, 2023.

\bibitem{gao2023enabling}
Tianyu Gao, Howard Yen, Jiatong Yu, and Danqi Chen.
\newblock Enabling large language models to generate text with citations.
\newblock In {\em Proceedings of the 2023 Conference on Empirical Methods in Natural Language Processing}, pages 6465--6488, 2023.

\bibitem{hatalis2023memory}
Kostas Hatalis, Despina Christou, Joshua Myers, Steven Jones, Keith Lambert, Adam Amos-Binks, Zohreh Dannenhauer, and Dustin Dannenhauer.
\newblock Memory matters: The need to improve long-term memory in llm-agents.
\newblock In {\em Proceedings of the AAAI Symposium Series}, volume~2, pages 277--280, 2023.

\bibitem{hongmetagpt}
Sirui Hong, Mingchen Zhuge, Jonathan Chen, Xiawu Zheng, Yuheng Cheng, Jinlin Wang, Ceyao Zhang, Zili Wang, Steven Ka~Shing Yau, Zijuan Lin, et~al.
\newblock Metagpt: Meta programming for a multi-agent collaborative framework.
\newblock In {\em The Twelfth International Conference on Learning Representations}.

\bibitem{izacard2020leveraging}
Gautier Izacard and Edouard Grave.
\newblock Leveraging passage retrieval with generative models for open domain question answering.
\newblock {\em arXiv preprint arXiv:2007.01282}, 2020.

\bibitem{jimenezswe}
Carlos~E Jimenez, John Yang, Alexander Wettig, Shunyu Yao, Kexin Pei, Ofir Press, and Karthik~R Narasimhan.
\newblock Swe-bench: Can language models resolve real-world github issues?
\newblock In {\em The Twelfth International Conference on Learning Representations}.

\bibitem{karpukhin2020dense}
Vladimir Karpukhin, Barlas O{\u{g}}uz, Sewon Min, Patrick Lewis, Ledell Wu, Sergey Edunov, Danqi Chen, and Wen-tau Yih.
\newblock Dense passage retrieval for open-domain question answering.
\newblock {\em arXiv preprint arXiv:2004.04906}, 2020.

\bibitem{kimi}
Kimi.ai.
\newblock Kimi.ai.
\newblock \url{https://www.perplexity.ai/}, 2023.
\newblock Accessed: 2025-01-09.

\bibitem{langchain}
LangChain.
\newblock Langchain.
\newblock \url{https://www.langchain.com/}, 2023.
\newblock Accessed: 2025-01-09.

\bibitem{li2023camel}
Guohao Li, Hasan Hammoud, Hani Itani, Dmitrii Khizbullin, and Bernard Ghanem.
\newblock Camel: Communicative agents for" mind" exploration of large language model society.
\newblock {\em Advances in Neural Information Processing Systems}, 36:51991--52008, 2023.

\bibitem{lin-etal-2024-arxiv}
Guanyu Lin, Tao Feng, Pengrui Han, Ge~Liu, and Jiaxuan You.
\newblock {A}rxiv copilot: A self-evolving and efficient {LLM} system for personalized academic assistance.
\newblock In Delia~Irazu Hernandez~Farias, Tom Hope, and Manling Li, editors, {\em Proceedings of the 2024 Conference on Empirical Methods in Natural Language Processing: System Demonstrations}, pages 122--130, Miami, Florida, USA, November 2024. Association for Computational Linguistics.

\bibitem{liu2023llm+}
Bo~Liu, Yuqian Jiang, Xiaohan Zhang, Qiang Liu, Shiqi Zhang, Joydeep Biswas, and Peter Stone.
\newblock Llm+ p: Empowering large language models with optimal planning proficiency.
\newblock {\em arXiv preprint arXiv:2304.11477}, 2023.

\bibitem{llamaindex}
LlamaIndex.
\newblock {LlamaIndex}: Build ai knowledge assistants over your enterprise data.
\newblock \url{https://www.llamaindex.ai/}, 2023.
\newblock Accessed: 2025-01-09.

\bibitem{lu2024chameleon}
Pan Lu, Baolin Peng, Hao Cheng, Michel Galley, Kai-Wei Chang, Ying~Nian Wu, Song-Chun Zhu, and Jianfeng Gao.
\newblock Chameleon: Plug-and-play compositional reasoning with large language models.
\newblock {\em Advances in Neural Information Processing Systems}, 36, 2024.

\bibitem{min2022rethinking}
Sewon Min, Xinxi Lyu, Ari Holtzman, Mikel Artetxe, Mike Lewis, Hannaneh Hajishirzi, and Luke Zettlemoyer.
\newblock Rethinking the role of demonstrations: What makes in-context learning work?
\newblock In {\em Proceedings of the 2022 Conference on Empirical Methods in Natural Language Processing}, pages 11048--11064, 2022.

\bibitem{parisi2022talm}
Aaron Parisi, Yao Zhao, and Noah Fiedel.
\newblock Talm: Tool augmented language models.
\newblock {\em arXiv preprint arXiv:2205.12255}, 2022.

\bibitem{perplexity}
Perplexity.
\newblock Perplexity.
\newblock \url{https://www.perplexity.ai/}, 2023.
\newblock Accessed: 2025-01-09.

\bibitem{ram2023context}
Ori Ram, Yoav Levine, Itay Dalmedigos, Dor Muhlgay, Amnon Shashua, Kevin Leyton-Brown, and Yoav Shoham.
\newblock In-context retrieval-augmented language models.
\newblock {\em Transactions of the Association for Computational Linguistics}, 11:1316--1331, 2023.

\bibitem{schick2023toolformer}
Timo Schick, Jane Dwivedi-Yu, Roberto Dess{\`\i}, Roberta Raileanu, Maria Lomeli, Eric Hambro, Luke Zettlemoyer, Nicola Cancedda, and Thomas Scialom.
\newblock Toolformer: Language models can teach themselves to use tools.
\newblock {\em Advances in Neural Information Processing Systems}, 36:68539--68551, 2023.

\bibitem{bing}
Bing Search.
\newblock Bing search.
\newblock \url{https://www.bing.com/}.
\newblock Accessed: 2025-01-09.

\bibitem{shinn2024reflexion}
Noah Shinn, Federico Cassano, Ashwin Gopinath, Karthik Narasimhan, and Shunyu Yao.
\newblock Reflexion: Language agents with verbal reinforcement learning.
\newblock {\em Advances in Neural Information Processing Systems}, 36, 2024.

\bibitem{tang2023medagents}
Xiangru Tang, Anni Zou, Zhuosheng Zhang, Ziming Li, Yilun Zhao, Xingyao Zhang, Arman Cohan, and Mark Gerstein.
\newblock Medagents: Large language models as collaborators for zero-shot medical reasoning.
\newblock {\em arXiv preprint arXiv:2311.10537}, 2023.

\bibitem{wei2022emergent}
Jason Wei, Yi~Tay, Rishi Bommasani, Colin Raffel, Barret Zoph, Sebastian Borgeaud, Dani Yogatama, Maarten Bosma, Denny Zhou, Donald Metzler, et~al.
\newblock Emergent abilities of large language models.
\newblock {\em arXiv preprint arXiv:2206.07682}, 2022.

\bibitem{wei2022chain}
Jason Wei, Xuezhi Wang, Dale Schuurmans, Maarten Bosma, Fei Xia, Ed~Chi, Quoc~V Le, Denny Zhou, et~al.
\newblock Chain-of-thought prompting elicits reasoning in large language models.
\newblock {\em Advances in neural information processing systems}, 35:24824--24837, 2022.

\bibitem{wei2022cot}
Jason Wei, Xuezhi Wang, Dale Schuurmans, Maarten Bosma, Fei Xia, Ed~Chi, Quoc~V Le, Denny Zhou, et~al.
\newblock Chain-of-thought prompting elicits reasoning in large language models.
\newblock {\em Advances in neural information processing systems}, 35:24824--24837, 2022.

\bibitem{agent_intro}
Lilian Weng.
\newblock {LLM} powered autonomous agents.
\newblock \url{https://lilianweng.github.io/posts/2023-06-23-agent/}, 2023.
\newblock Accessed: 2025-01-09.

\bibitem{feijie}
Feijie Wu, Zitao Li, Fei Wei, Yaliang Li, Bolin Ding, and Jing Gao.
\newblock Talk to right specialists: Routing and planning in multi-agent system for question answering, 2025.

\bibitem{wu2023autogen}
Qingyun Wu, Gagan Bansal, Jieyu Zhang, Yiran Wu, Shaokun Zhang, Erkang Zhu, Beibin Li, Li~Jiang, Xiaoyun Zhang, and Chi Wang.
\newblock Autogen: Enabling next-gen llm applications via multi-agent conversation framework.
\newblock {\em arXiv preprint arXiv:2308.08155}, 2023.

\bibitem{yang2024swe}
John Yang, Carlos~E Jimenez, Alexander Wettig, Kilian Lieret, Shunyu Yao, Karthik Narasimhan, and Ofir Press.
\newblock Swe-agent: Agent-computer interfaces enable automated software engineering.
\newblock {\em arXiv preprint arXiv:2405.15793}, 2024.

\bibitem{yao2024tree}
Shunyu Yao, Dian Yu, Jeffrey Zhao, Izhak Shafran, Tom Griffiths, Yuan Cao, and Karthik Narasimhan.
\newblock Tree of thoughts: Deliberate problem solving with large language models.
\newblock {\em Advances in Neural Information Processing Systems}, 36, 2024.

\bibitem{yao2022react}
Shunyu Yao, Jeffrey Zhao, Dian Yu, Nan Du, Izhak Shafran, Karthik Narasimhan, and Yuan Cao.
\newblock React: Synergizing reasoning and acting in language models.
\newblock {\em arXiv preprint arXiv:2210.03629}, 2022.

\end{thebibliography}

\end{document}